\documentclass{article}

\usepackage{amsmath}
\usepackage{algorithmic}
\usepackage{algorithm}
\usepackage{placeins}
\usepackage{bbold}

\DeclareMathOperator*{\argmax}{argmax}

\newcommand{\gr}{\mathcal{G}}
\newcommand{\p}[1]{#1^{\prime}}

\newcommand{\states}{\mathcal{S}}
\newcommand{\actions}{\mathcal{A}}

\usepackage[short]{optidef}

\newcommand{\E}{\mathbb{E}}

\newcommand{\g}{\mathcal{G}}
\newcommand{\x}{\mathcal{X}}

\newcommand{\s}{\mathcal{S}}
\newcommand{\A}{\mathcal{A}}

\usepackage[final]{optrl_2019}

\usepackage[utf8]{inputenc} %
\usepackage[T1]{fontenc}    %
\usepackage{hyperref}       %
\usepackage{amsthm}
\usepackage{url}            %
\usepackage{booktabs}       %
\usepackage{amsfonts}       %
\usepackage{nicefrac}       %
\usepackage{microtype}      %
\usepackage{mathtools}
\usepackage{authblk}
\usepackage{lmodern}

\mathtoolsset{showonlyrefs=true}

\title{On Connections between Constrained Optimization and Reinforcement Learning}

\author{Nino Vieillard}
\author{Olivier Pietquin}
\author{Matthieu Geist}
\affil{Google Research, Brain Team}
\date{}

\begin{document}
\maketitle

\begin{abstract}
    Dynamic Programming (DP) provides standard algorithms to solve Markov Decision Processes. However, these algorithms generally do not optimize a scalar objective function. In this paper, we draw connections between DP and (constrained) convex optimization. Specifically, we show clear links in the algorithmic structure between three DP schemes and optimization algorithms. We link conservative Policy iteration to Frank-Wolfe, Mirror-Descent Modified Policy Iteration to Mirror Descent, and Politex (Policy Iteration Using Expert Prediction) to Dual Averaging. These abstract DP schemes are representative of a number of (deep) reinforcement learning (RL) algorithms. By highlighting these connections (most of which have been noticed earlier, but in a scattered way), we would like to encourage further studies linking RL and convex optimization, that could lead to the design of new, more efficient, and better understood RL algorithms.
\end{abstract}

\section{Introduction}

A standard way to model sequential decision making is through the frame of Markov Decision problems (MDPs). These MDPs, when known and small enough, can be solved with Dynamic Programming (DP), of which two classical algorithms are policy iteration (PI) and value iteration (VI). Solving an MDP means finding a policy $\pi_*$ such that its value $v_{\pi_*}$ dominates the value of any other policy, state-wise. As such, it cannot be easily framed as a (convex) optimization problem, even if connections can be made: VI can be (indirectly) seen as minimizing the residual $\|T_* v - v\|$ through a fixed-point approach, and PI can be obtained as Newton descent on the same residual (\textit{e.g.},~\citep{pmlr-v48-perolat16} and references therein).

Reinforcement learning aims at solving sequential decision making problems through interactions. Many RL algorithms can be seen as approximate and possibly asynchronous and stochastic variations of DP algorithms, mainly PI and VI. As such, connections of these algorithms to convex optimization are quite loose. A notable exception is algorithms based on policy gradient, that aim at maximizing the expected value function, $J(\pi) = \E_{s\sim\mu}[v_\pi(s)]$, or a proxy of this objective function~\citep{deisenroth2013survey}. However, this function is not concave, which is often a problem from an optimization perspective. Another exception is the set of algorithms that aim at minimizing the residual $J(v) = \|T_* v - v\|$. Again, this is not convex in general and this approach comes with drawbacks~\citep{geist2017bellman}.

In this work, we describe three connections between DP schemes and convex optimization algorithms, from an algorithmic perspective, and without considering estimation or approximation errors. We relate the Frank-Wolfe algorithm~\citep{frank1956algorithm} to Conservative Policy Iteration~\citep{kakade2002approximately} in Section~\ref{sec:cpi-fw}. Then, we establish a connection between Mirror Descent~\citep{beck2003mirror} and Mirror Descent Modified Policy Iteration~\citep{geist2019theory} in Section~\ref{sec:md-md}, and finally between Dual Averaging~\citep{nesterov2009primal} and Politex~\citep{lazic2019politex} in Section~\ref{sec:politex-da}. Most of these connections have been noticed earlier, but in a scattered way, and sometimes not clearly. We also discuss briefly more connections between RL and optimization in Section~\ref{sec:other-connections}.

\section{Background}

\subsection{Markov Decision Processes}

We consider the problem of solving infinite horizon discounted Markovian Decision Process (MDP). An MDP is a tuple $\{\states, \actions, P, r, \gamma\}$, where $\states$ and $\actions$ are finite spaces of respectively states and actions, $P \in\Delta_\states^{\states\times\actions}$ is a Markovian transition kernel (writing $\Delta_X$ the simplex over the set $X$), $r\in  \mathbb{R}^{\states\times\actions}$ a reward function and $\gamma \in (0,1)$ the discount factor. The objective of dynamic programming is to find a policy $\pi_*$ that dominates every policy $\pi$ in the space of policies that we denote $\Pi$. A policy $\pi \in \Pi$ is a function that associate to each state $s\in \states$ a distribution over $\actions$,  $\pi(\cdot | s)$. It defines an associated stochastic kernel, $P_\pi(\p s | s) = \mathbb{E}_{a \sim \pi(\cdot | s)}[P(\p{s}|s,a)]$, and an associated expected reward function $r_\pi(s) = \mathbb{E}_{a \sim \pi(\cdot | s)}[r(s,a)]$. The value $v_\pi \in \mathbb{R}^{\states}$ of a policy associates to each state the expected discounted cumulative reward,
\begin{equation}
    v_\pi(s) = \mathbb{E}\left[\sum_{t=0}^{\infty} \gamma^tr(s_t, a_t) \,\middle\vert\, s_0=s, s_{t+1} \sim P(\cdot | s_t, a_t), a_{t} \sim \pi(\cdot | s_{t}) \right].
\end{equation}

To every policy is associated a Bellman evaluation operator, defined for each $v \in \mathbb{R}^{\states}$ as $T_\pi v = r_\pi + \gamma P_\pi v$. The value function of a policy is the unique fixed point of this operator. The Bellman optimality operator is defined as $T_* v = \max_\pi T_\pi v$, and the optimal value function $v_*$ verifies $v_* = T_* v_*$. We define the set of greedy policies w.r.t. to any value $v$ as $\gr v = \{ \pi \vert T_\pi v = T_* v \}$. An optimal policy $\pi_*$ is such that $\pi_* \in \gr v_*$.

The state-action value function $q_\pi \in\mathbb{R}^{\states \times \actions}$ associated to a policy $\pi$ is defined as
\begin{equation*}
    q_\pi(s,a) = r(s,a) + \gamma \mathbb{E}_{\p{s} \sim P(\cdot|s,a)}[v_\pi(\p{s})].
\end{equation*}
This state-action value function possesses similar properties to the value function, In particular, we can easily re-write the Bellman equations using state-action values instead of values, and we have that $T_\pi q_\pi = q_\pi$ and $T_* q_{\pi_*} = q_{\pi_*} = q_*$. The greediness can be defined w.r.t to a state-action value by $\gr q = \argmax_a q(\cdot,a)$.  

Two classic DP algorithms are PI and VI. PI iterates as follows, $\pi_{k+1}\in\g(v_{\pi_k})$, its convergence being guaranteed by the fact that for any policy $\pi$, $v_{\gr v_\pi} \geq v_{\pi}$, the equality being obtained only at optimality. VI iterates as follows, $v_{k+1} = T_* v_k$, its convergence being guaranteed by the fact that the Bellman operator is a $\gamma$-contraction. In both cases, the argument for convergences are loosely related to convex optimization. Both schemes can be generalized to modified policy iteration (MPI), that iterates as
\begin{equation}
\begin{cases}
    v_{k} = T_{\pi_{k}}^m v_{k-1} \\ \pi_{k+1} \in \gr v_k .
\end{cases}
\end{equation}
We retrieve PI with $m=\infty$ (in this case, we compute the fixed point of the contractive operator $T_{\pi_k}$) and VI with $m=1$ (noticing that $v_k = T_{\pi_{k}} v_{k-1} = T_{\gr v_{k-1}} v_{k-1} = T_* v_{k-1}$). This can be seen as a PI scheme where the full evaluation of the policy is relaxed to a partial, less costly, evaluation.

We define the objective function $J \in \mathbb{R}^{\Delta_\actions^\states}$ as
\begin{equation}
    \label{eq:objective}
    J(\pi) = \mathbb{E_{s \sim \mu}}\left[v_{\pi}(s)\right],
\end{equation}
with $\mu$ a given distribution of states. We aim at solving $\max_{\pi \in \Delta_\actions^\states} J(\pi)$. This non-concave optimization problem can be addressed for example with a gradient ascent, and it is the basis of a number of RL algorithms~\citep{deisenroth2013survey}, either directly or indirectly. For example, Conservative Policy Iteration~\citep{kakade2002approximately} or Trust Region Policy Optimization (TRPO)~\citep{schulman2015trust} were designed for optimizing a proxy to this objective function. An interesting observation is that the natural gradient of $J(\pi)$, $\tilde{\nabla} J(\pi)$, in the tabular case (and with a softmax representation for the policy), is indeed the state-action value function, that is $\tilde{\nabla} J(\pi) = q_\pi$. This is a direct corollary of the results of~\cite{kakade2002natural}. We'll use this observation later.

\subsection{Convex Optimization}
We study the problem of finding the maximum of a concave function over a convex set, instead of the standard setting where one aims to find the minimum of a convex function. This allows us to draw more natural connections to RL, without loss of generality.  More precisely, given a concave function $f$ taking its values in $\mathbb{R}$ and a closed convex space $\mathcal{X}$, we define a maximization problem as finding $x_* \in\mathcal{X}$ such that
\begin{equation}
    f(x_*) = \max_{x \in \mathcal{X}} f(x).\label{eq:concav_constrained}
\end{equation}

We consider methods derived from the gradient ascent scheme. This classic algorithm assumes that $f$ is (sub)differentiable, and produces a sequence $(x_k)$, with $\eta$ a learning rate, 
\begin{equation}
    x_{k+1} = x_k + \eta \nabla f(x_k).
\end{equation}
There is no reason for this sequence to stay in $\mathcal{X}$, and a classical solution is projected gradient ascent, that additionally project back the iterate onto the convex set,
\begin{equation}
\begin{cases}
    y_{k+1} = x_k + \eta \nabla f(x_k) \\
    x_{k+1} = \Phi_{\mathcal{X}}(y_{k+1}),
\end{cases}
\end{equation}
with $\Phi_{\mathcal{X}}$ the projection onto $\mathcal{X}$. We'll discuss other algorithms while drawing connections to RL in the following sections.

\section{Conservative Policy Iteration and Frank Wolfe}
\label{sec:cpi-fw}

We start by linking the classic Frank Wolfe algorithm~\citep{frank1956algorithm}, also known as Conditional Gradient Descent, to the DP scheme Conservative Policy Iteration (CPI,~\cite{kakade2002approximately}). This link was first made by~\citet{scherrer2014local}, who studied a boosting generalization of CPI based on Frank-Wolfe, but without mentioning it, and was explicitly  given by~\citet{shani2019adaptive}.

\paragraph{Frank--Wolfe} Frank-Wolfe (FW) addresses problem~\eqref{eq:concav_constrained}. Given the current iterate $x_k$ and its gradient $g_k = \nabla f(x_k)$, it searches for the element of $\mathcal{X}$ being the most colinear to the gradient, and update the iterate by doing a convex mixture of these elements, which ensure that the iterates stay in the convex set. Formally, it iterates as follows, with $x_0\in \x$ for initialization:
\begin{equation}
    \begin{cases}
    g_k = \nabla f(x_k)
    \\
    s_k = \argmax_{x\in \x} \langle x, g_k\rangle
    \\
    x_{k+1} = (1-\alpha) x_k + \alpha s_k.
    \end{cases}\label{eq:fw-scheme}
\end{equation}
Compared to projected gradient ascent, it might be easier in some problems to find the most colinear element than to project the iterate onto the convex set.

\paragraph{Conservative Policy Iteration} 
CPI is a classic RL algorithms that soften the PI scheme by replacing the greedy policy by a stochastic mixture of the greedy policy and the previous one. It iterates as follows:
\begin{equation}
    \pi_{k+1} = (1-\alpha) \pi_k + \alpha \g q_{\pi_{k}},\label{eq:cpi}
\end{equation}
with $\g q_{\pi_{k}}$ any greedy policy respectively to $q_{\pi_{k}}$. Without approximation, the optimal rate is $\alpha=1$, and the algorithm is of practical interest when there is approximation, but we stick to the exact case, the aim being to draw connections to optimization. 

Let $\mu\in\Delta_\s>0$ be a distribution over states\footnote{Notice that the original CPI considers the distribution $d_{\mu,\pi_k}=(1-\gamma)\mu(I-\gamma P_{\pi_k})^{-1}$ (the $\gamma$-weighted occupancy measure induced by $\pi_k$ for the initial state distribution $\mu$) instead of $\mu$, which is of importance for the analysis in the approximate setting, but does not change the algorithm in the exact setting considered here.\label{ftn:1}}, and define the scalar product on $\mathbb{R}^{\s\times\A}$ as $\langle q_1, q_2\rangle = \sum_s \mu(s) \sum_a q_1(s,a) q_2(s,a)$. The greedy policy $\gr q_{\pi_{k}}$ can be equivalently seen as the policy maximizing $\langle \pi, q_{\pi_{k}}\rangle$. Recall also that the state-action value function is indeed the natural gradient of $J$, $\tilde{\nabla} J(\pi) = q_\pi$. Given this, we can rewrite the update of CPI as follows:
\begin{equation}
    \begin{cases}
        q_k = q_{\pi_k} = \tilde{\nabla} J(\pi_k)
        \\
        \pi'_k = \argmax_{\pi\in\Delta_\A^\s} \langle \pi, q_k\rangle
        \\
        \pi_{k+1} = (1-\alpha) \pi_k + \alpha \pi'_k.
    \end{cases} \label{eq:cpi-scheme}
\end{equation}
Comparing Eqs.~\eqref{eq:fw-scheme} and~\eqref{eq:cpi-scheme}, it is clear that CPI is indeed the Frank-Wolfe algorithm applied to the objective function $J(\pi)$, up to the fact that the vanilla gradient is replaced by the natural gradient\footnote{With the vanilla gradient, the derivation would be a bit more invovled, but we would obtain the original CPI, that is with $\mu$ being replaced by $d_{\mu,\pi_k}$, see the derivations of~\citet{scherrer2014local}.\label{ftn:2}}. We are not aware of a FW algorithm considering natural gradients in the optimization literature.

CPI being a variation of the PI scheme, we can also write an MPI-like variation of CPI, as considered by~\citet{vieillard2019deep}. This gives the following iterate (with $\pi_0$ and $q_0$ for initialization), that generalizes scheme~\eqref{eq:cpi-scheme}:
\begin{equation}
    \begin{cases}
        q_k = T_{\pi_k}^m q_{k-1}
        \\
        \pi'_k = \argmax_{\pi\in\Delta_\A^\s} \langle \pi, q_k\rangle
        \\
        \pi_{k+1} = (1-\alpha) \pi_k + \alpha \pi'_k.
    \end{cases}
\end{equation}
With $m<\infty$, we no longer have a FW scheme, but the algorithmic approach is very similar. In addition to the process updating the policies, we have an additional process that replaces the gradient computation, this process iterating value functions based on the Bellman operator of the current policy. With this point of view, the  function $q_k$ plays a role similar to the gradient, giving the direction of improvement. Notice that this scheme converges, despite the lack of concavity~\citep{vieillard2019deep} (but the proof does not rely on optimization-like arguments).

\section{MD-MPI and Mirror Descent}
\label{sec:md-md}
Here, we show a connection between the Mirror Descent (MD,~\citet{beck2003mirror}) optimization method and the DP scheme Mirror Descent Modified Policy iteration (MD-MPI,~\citet{geist2019theory}). 

\paragraph{Mirror Descent} Let $D_\Omega$ be the Bregman divergence generated by a convex function $\Omega \in \mathbb{R}^\x$, $D_\Omega(x||x') = \Omega(x) - \Omega(x') - \langle \nabla\Omega(x'), x - x'\rangle$. Classic Bregman divergences are the Euclidean distance, generated by the $\ell_2$ norm, or the Kullback-Leibler (KL) divergence, generated by the negative entropy. Mirror Descent solves a maximization problem by iterating as follows:
\begin{equation}
    \begin{cases}
        g_k = \nabla f(x_k)
        \\
        x_{k+1} = \argmax_{x\in \x} \eta \langle x, g_k\rangle - D_\Omega(x||x_k).
    \end{cases}\label{eq:md-scheme}
\end{equation}
In words, MD consists at each step in maximizing a linearization of the function of interest, with the constraint of not moving to far from the previous iterate, this constraint being quantified by the Bregman divergence. It generalizes classic gradient ascent, that can be retrieved in the unconstrained case by considering the Euclidean distance.

\paragraph{MD-MPI} \citet{geist2019theory} introduced a set of abstract DP algorithms to study the propagation of errors of regularized RL algorithms, where the greediness is softened by a convex regularizer (typically, a scaled entropy or KL divergence, see the paper for details). MD-MPI considers the general case where the greediness is regularized by a Bregman divergence, within an MPI-like scheme. We consider a slight variation here, where greediness is computed in expectation with respect to $\mu$ (instead of state-wise in the original work).

Let $\Omega:\Delta_\A\rightarrow\mathbb{R}$ be a convex function, and define (with a slight abuse of notation) the convex function $\Omega:\Delta^\s_\A\rightarrow \mathbb{R}$ as $\Omega(\pi) = \sum_s \mu(s) \Omega(\pi(.|s))$. 
We first consider a PI-like update of MD-MPI (that is, setting $m=\infty$, or MD-PI), and recall that $q_{\pi_k} = \tilde{\nabla} J(\pi_k)$. The corresponding update rule can be written as
\begin{equation}
    \begin{cases}
        q_k = q_{\pi_k} = \tilde{\nabla} J(\pi_k)
        \\
        \pi_{k+1} = \argmax_{\pi\in \Delta_\s^\A} \eta \langle \pi, q_k\rangle - D_\Omega(\pi||\pi_k).
    \end{cases}\label{eq:md-mpi-scheme}
\end{equation}
As before, comparing Eqs.~\eqref{eq:md-scheme} and~\eqref{eq:md-mpi-scheme}, MD-PI is exactly mirror descent applied to $J(\pi)$, up to the fact that the vanilla gradient is replaced by the natural one. We're not aware of any MD algorithm doing so in the optimization literature.

With a KL divergence, this scheme is very close to TRPO, the main differences being that the distribution $d_{\mu,\pi_{k}}$ (defined in footnote~\ref{ftn:1}) is used instead of $\mu$, at each iteration. Contrary to CPI, that can be seen exactly as Frank-Wolfe with vanilla gradient (see footnote~\ref{ftn:2}), Eq.~\eqref{eq:md-mpi-scheme} with the vanilla gradient would not give TRPO. The reader can refer to~\citet{shani2019adaptive}, who draw connections between MD and TRPO, and show that TRPO is a kind of modified MD.

We can also write the original MD-MPI (up to the slight variation mentioned before):
\begin{equation}
    \begin{cases}
        q_k = T_{\pi_k}^m q_{k-1}
        \\
        \pi_{k+1} = \argmax_{\pi\in \Delta_\s^\A} \eta \langle \pi, q_k\rangle - D_\Omega(\pi||\pi_k)
    \end{cases}.
\end{equation}
As it was the case for CPI, we obtain a mirror descent scheme, up to the fact that the gradient is replaced by a second iterative process on the $q$-values. In both domains, the idea is similar: the goal is to move in the direction of improvement -- either a new point or a new policy --, this direction being provided by the gradient or the $q$-value, but staying not too far from the previous point or policy.

\section{Politex and Dual Averaging}
\label{sec:politex-da}

Finally, we show a new connection between the recent Politex (Policy Iteration Using Expert Prediction)  algorithm~\citep{lazic2019politex}, and the Dual Averaging~(DA, \citet{nesterov2009primal}) optimization scheme.  

\paragraph{Dual Averaging} Dual Averaging can be considered as a lazy version of MD (see~\citet[Section 4.4]{bubeck2015convex}), and iterates as follows
\begin{equation}
    \begin{cases}
        g_k = \nabla f(x_k)
        \\
        x_{k+1} = \argmax_{x\in \x} \eta \langle x, \sum_{j=0}^k g_j\rangle - \Omega(x).
    \end{cases}\label{eq:da-scheme}
\end{equation}
In some cases, it can be shown to be equivalent to MD. This algorithmic scheme can also be linked to Follow the Regularized Leader (FTRL), that is somehow a building block of Politex (the related paper focusing indeed on prediction with expert advice, which is related).

\paragraph{Politex} Politex is a PI scheme, in the average reward case, that, at each iteration, estimates the value of its current policy, and computes a new policy that is the softmax of the sum of all $q$-values of the preceding policies. Notice that a softmax policy is a greedy policy regularized by an entropy (\textit{e.g.}, see~\citet{geist2019theory}).
Here, we first consider this algorithm in the cumulative discounted reward case, generalized to any regularizer. In this case, we can write a generalization of Politex as 
\begin{equation}
    \begin{cases}
        q_k = q_{\pi_k} = \tilde{\nabla} J(\pi_k)
        \\
        \pi_{k+1} = \argmax_{\pi\in \Delta_\s^\A} \eta \langle \pi, \sum_{j=0}^k q_j\rangle - \Omega(\pi)
    \end{cases}.\label{eq:politex-scheme}
\end{equation}
Comparing Eqs.~\eqref{eq:da-scheme} and~\eqref{eq:politex-scheme}, we see that this variation of Politex is exactly DA with a natural gradient in lieu of the vanilla one.

This can be easily generalized to an MPI-like scheme:
\begin{equation}
    \begin{cases}
        q_k = T_{\pi_k}^m q_{k-1}
        \\
        \pi_{k+1} = \argmax_{\pi\in \Delta_\s^\A} \eta \langle \pi, \sum_{j=0}^k q_j\rangle - \Omega(\pi).
    \end{cases}
\end{equation}
Again, this is DA, up to the fact that the gradient is replaced by an iterative process for computing $q$-values.

\section{Other connections}
\label{sec:other-connections}

There are other works, not discussed above, that draw connections between optimization and RL. We mention briefly some of them here.

\citet{bertsekas1996neuro} see $v - T_*v$ as the gradient of an unknown function, and this has served as a building block to~\citet{goyal2019first} to introduce a Nesterov-like VI algorithm, with improved convergence rate.

Another classic DP approach, less used in RL, compute the optimal value function as the solution of a linear program. \citet{neu2017unified} build on this to study regularized MDPs, where optimal policies are computed as the solution of real convex problems. The practical issue is that it involves working in the dual space (on state-action distributions), which might be not practical (as required quantities might not be easy to estimate).

\citet{agarwal2019optimality} study the convergence of various policy gradient algorithms, and despite the lack of concavity, their proof techniques borrow to optimization (and allow showing even global convergence in some cases).

\section{Discussion}
\label{sec:discussion}

Using an analogy between the gradient and the state-action value function, we have shown a strong connection between some constrained convex optimization and RL algorithms. We have made this connection clear on three cases: Frank Wolfe and Conservative policy iteration, Mirror Descent and MD Modified Policy Iteration, and Dual Averaging and Politex. A summary of these connections is presented in Table~\ref{tab:summary}. All these algorithms are improvements of canonical algorithms in their fields, namely Projected Gradient Descent in optimization and policy iteration in dynamic programming. They all share a same idea of regularization, either by a stochastic mixture of consecutive policies (like in CPI), or by explicitly regularizing the improvement step (like in mirror descent, that regularizes its improvement with a divergence).

From these three examples, we see how some methods have been identified in Optimization and in RL, and this sheds light and links that could be exploited in RL. Indeed, such a connection could be useful for either designing new algorithms in RL, or giving new techniques of analysis to better understand these algorithms. Drawing more formal links between optimization and RL (for example, linking the contraction in RL to strong convexity or Lipschitzness in optimization) could also be useful to strengthen the general connection between these fields.

\begin{table}[]
    \centering
    \caption{Summary of connections and analogies.}
    \begin{tabular}{ll}
        \toprule
         Convex Optimization & Reinforcement Learning  \\
         \midrule
         \hspace{3pt}$x$ & $\pi$ \\
         $f(x)$ & ? ($J(\omega)$ in tabular case) \\
         $\nabla f(x)$ & $q_\pi$ \\
         \midrule
         Frank--Wolfe & Conservative Policy Iteration \\
         Mirror Descent & MD-MPI (type 2) \\
         Dual Averaging & Politex \\
         \bottomrule
    \end{tabular}
    \label{tab:summary}
\end{table}

\clearpage

\bibliographystyle{plainnat}
\bibliography{biblio}

\end{document}